\documentclass[journal,twocolumn]{IEEEtran}
\ifCLASSINFOpdf
\else
\fi
\usepackage{graphicx}
\usepackage{subcaption}
\usepackage{booktabs}
\usepackage{caption}

\begin{document}
%
\title{Towards automatic smoke detector inspection: Recognition of the smoke detectors in industrial facilities and preparation for future drone integration}
%
%
%

\author{Lukas~Kratochvila,
        Jakub~Stefansky,
        Simon~Bilik,
        Robert~Rous,
        Tomas~Zemcik,
        Michal~Wolny,
        Frantisek~Rusnak,
        Ondrej~Cech,
        and~Karel~Horak
\thanks{L. Kratochvila, T. Zemcik, F. Rusnak, and K. Horak are with the Department of Control and Instrumentation, Brno University of Technology, Brno, Czech Republic. E-mail: kratochvila@vut.cz.}
\thanks{J. Stefansky is with the Department of Cybernetics and Biomedical Engineering, VSB - Technical University of Ostrava, Ostrava, Czech Republic.}
\thanks{S. Bilik is with Institute for Research and Applications of Fuzzy Modeling, University of Ostrava, Ostrava, Czech Republic and with Department of Informatics, Mendel University in Brno, Brno, Czech Republic. E-mail: simon.bilik@osu.cz}
\thanks{R. Rous is with Department of Informatics, Mendel University in Brno, Brno, Czech Republic.}
\thanks{O. Cech is with Department of Electrical and Electronic Technology, Brno University of Technology, Brno, Czech Republic}
}

%
%

\markboth{Preprint submitted to ArXiv}%
{Shell \MakeLowercase{\textit{et al.}}: Towards automatic smoke detector inspection: Recognition of the smoke detectors in industrial facilities and preparation for future drone integration}
%



\maketitle

\begin{abstract}
Fire safety consists of a complex pipeline, and it is a very important topic of concern. One of its frontal parts are the smoke detectors, which are supposed to provide an alarm prior to a massive fire appears. As they are often difficult to reach due to high ceilings or problematic locations, an automatic inspection system would be very beneficial as it could allow faster revisions, prevent workers from dangerous work in heights, and make the whole process cheaper. In this study, we present the smoke detector recognition part of the automatic inspection system, which could easily be integrated to the drone system. As part of our research, we compare two popular convolutional-based object detectors YOLOv11 and SSD widely used on embedded devices together with the state-of-the-art transformer-based RT-DETRv2 with the backbones of different sizes. Due to a complicated way of collecting a sufficient amount of data for training in the real-world environment, we also compare several training strategies using the real and semi-synthetic data together with various augmentation methods. To achieve a robust testing, all models were evaluated on two test datasets with an expected and difficult appearance of the smoke detectors including motion blur, small resolution, or not complete objects. The best performing detector is the YOLOv11n, which reaches the average mAP@0.5 score of 0.884. Our code, pretrained models and dataset are publicly available.
\end{abstract}


%
\IEEEpeerreviewmaketitle

\section{Introduction}
%
%
%
%
\IEEEPARstart{S}{moke} sensors are one of the crucial parts of fire prevention equipment and are now more widespread than ever - even mandatory for certain building types in most countries. The tangible effects of their installation can be seen in a significant reduction of fatalities per fire occurrence when installed in homes -~\cite{McGree2024} reports up to 50 \% reduction in fatalities in US homes equipped with smoke alarms in 2024. ~\cite{Festag2026} reports similarly significant improvements observed over the 1998-2022 period in all the 16 German federal states including the period before, during and after the obligatory smoke alarm installation. Smoke sensors have proven themselves a working and useful tool, that can only remain effective if properly maintained and tested.

In industrial and storage facilities, the inspection process might be complicated due to high ceilings or difficult to reach locations. Therefore, we present a robust detection framework designed to be used as part of a complex drone inspection system for periodic checks of smoke sensors using a drone carrying a camera and a dedicated sensor-testing device. Such a system has the potential to replace dangerous and laborious checks done by human workers, who have to work under difficult conditions at heights, and due to its simplicity also allows more frequent checks to increase the safety of these facilities. Potential field of use is, however not limited only to this application, as the presented approaches and code could easily be used in other similar domains as well.

A serious problem to be concerned with when focusing on this task is the collection of a sufficient amount of data to train deep learning models. Such a problem has a similar cause as a manual sensor check - sensors are difficult to reach, often placed in heights, and the industrial areas are often restricted. Therefore, some of the methods to augment existing~\cite{buslaev2020albumentations} or generate new synthetic~\cite{man2022review, koetzier2024generating} or semi-synthetic data~\cite{dondi2025post} should be deployed to train the selected object detector architectures.

In this study, we focus on single-stage object detectors such as SSD~\cite{10.1007/978-3-319-46448-0_2}, or YOLO~\cite{redmon2016you} as they generally achieve a lower inference times and energy consumption in comparison with the two-stage approach, which balance their lower accuracy as suggested for example in~\cite{Dai_2020}. Both facts are important from the perspective of using the trained model on a drone with limited battery capacities. Nevertheless, implementations of the two-stage object detectors exist as well, for example the Faster R-CNN~\cite{fasterRCNN}. To complement our experiments, we decided to use a state-of-the-art object detector based on transformers - RT-DETRv2~\cite{lv2024rtdetrv2improvedbaselinebagoffreebies} although its higher demands on computational hardware especially during training and worse inference times in comparison with the YOLO object detector family limit their usage potential. Finally, we also focus on methods to deploy the best performing model to the drone.

\bigskip

\subsection{Research questions}

Our research questions are set as follows:

\begin{itemize}
    \item How could an existing dataset be efficiently augmented and what is the optimal training strategy for using synthetic data?
    \item Which configuration of existing object detectors is the most suitable for smoke sensor detection?
    \item What effect does custom augmentation have on the selected object detector and how do the detection results differ with a unified augmentation technique?
    \item How could an object detector be utilized for use in an embedded device?
\end{itemize}


\subsection{Novelty and contributions}

The main contribution of this study is its orientation towards deployment in a visual industrial application with a limited dataset and data-efficient learning. We compare three object detectors: the convolution-based state-of-the-art YOLOv11 and legacy SSD together with the transformer-based state-of-the-art RT-DETRv2. For a practical deployment, we also discuss the licensing issues associated with using these detectors.

As assembling a dataset of images of fire sensors in real environments is difficult due to their limited reachability in heights, we explore the possibility of using rendered sensor images and captures of real sensors in a laboratory environment in combination with real backgrounds to create a semi-realistic dataset to be used for training or validation. Such datasets are used in five different training strategies and the results are evaluated over two test datasets containing images with normal and difficult appearance.

All three detectors are evaluated in several variants. As an out of distribution test, we compare the best performing models using custom augmentations simulating conditions such as defocus or motion blur which might occur while deployed on the drone. We also performed several additional experiments focusing on the most optimal YOLOv11 detectors, which are described in the discussion section.

Finally, we also present a custom ROS-2 pipeline for future integration into the detection system together with an analysis and discussion of the inference times achieved on two versions of RaspberryPi embedded computers and other parameters important for a deployment on an embedded platform.




\section{Object detection on the edge devices}
\label{sec:objectdetembb}

For the aforementioned architectures to be applicable to the problem at hand, the models need to be readily deployable on a highly mobile platform which is power, mass, and volume constrained. This severely limits the number of available computational options to only the most compact of edge devices. Modern methods exist to overcome these limitations as outlined in surveys such as~\cite{luo2025efficientdeeplearninginfrastructures} on general deep learning models for embedded devices, or~\cite{millar2025energyawaredeeplearningresourceconstrained} on energy-aware deep learning architectures.

Object detection on the edge devices has become popular recently because of the wide availability of the one-board computers such as RaspberryPi, or NVIDIA Jetson families together with the emergence of more optimized deep learning techniques, which allow an efficient solution of relatively complex tasks such as the facial recognition, traffic sign detection, and other various home assistants in real-time with reduced latency and no computation occurring on remote servers (with associated increased privacy and security risks). The greatest limitations of this approach are especially limited computational resources, energy consumption, and model size  and complexity. In most areas, conventional computer vision techniques are outperformed by deep learning approaches~\cite{10374363}.

The authors of~\cite{10374363} further describe trends in the object detection on edge devices focusing on the methods for network optimization as distillation, pruning, and quantization together with investigating the balance between parameter reduction and desired accuracy. The authors also provide an overview of the most common edge computational devices with their parameters and they also compare the existing surveys with their strengths, contributions and limitations.

A number of studies have been conducted investigating the performance of various deep learning based object detectors on edge devices, ~\cite{alqahtani2024benchmarkingdeeplearningmodels} compared common object detectors, namely YOLOv8, SSD and EfficientDet Lite on available and accessible edge devices, several versions of the Raspberry Pi and Nvidia Jetson were used with and without TPU accelerators. The study looked at the basic tradeoff between performance (using mAP as the metric), inference times and energy consumption and highlighted some issues with converting models to run using the accelerators.

In ~\cite{machines13080684} object detectors (mostly from the YOLO family) were trained on the DroneVis dataset ~\cite{VisDrone2021} and evaluated on Nvidia Jetson Nano Super and Iris Xe Graphics devices, the study found that on given hardware some of the models run sub 25ms inference times for a given resolution of 960x960 and can thus be considered real-time, versions of YOLOv12 also outperformed some transformer based solutions such as RT-DETR. Differences were also found in inference times of certain methods such as YOLOv11 vs YOLOv12 which do not manifest on non-edge devices showing that some variants are more suited to edge computing.

In the article~\cite{10.1145/3301326.3301369}, the authors propose a system involving a RaspberryPi 3 with the Neural Compute Stick (NCS) accelerator and a remote server used for training of the SSD network with the MobileNet backbone. The object detector was trained using the Pascal VOC~\cite{everingham2010pascal} and KITTI~\cite{Geiger2013IJRR} datasets. For the evaluation, an own dataset called LPS2017 was built. The achieved inference time was 1.7 second using one NCS and 10 ms using four NCS accelerators. Reported achieved accuracy was around 63 \%.

The authors of ~\cite{Dai_2020} compare Faster-RCNN and SSD object detectors on the MS COCO dataset ~\cite{lin2015microsoft} while achieving 58 FPS with mAP 72.1 \% using the SSD300 model, 23 FPS with mAP 75.1 \% using the SSD500 model and 7 FPS with 73.2\% mAP using the Faster R-CNN. Analyzing these values, the authors propose further optimization of the SSD model using TensorFlow lite to allow its use on the edge devices.

In the paper~\cite{9219319}, the authors introduce a novel object detector using the MobileNetv2 backbone. The detector is designed to be used in ADAS systems and it was successfully implemented on the NVIDIA Jetson Xavier platform. The authors extract different feature maps and include inverse residual modules for an additional feature map extraction. The proposed network achieved approximately 5 \% higher mAP and a 15 \% higher FPS in comparison with the MobileNetv2 backboned SSD.

The study~\cite{9604698} focuses on recognition of possible threats using a drone equipped with a camera and RaspberryPi 4B computer. Authors compare the performance of Faster-RCNN and SSD MobileNet V2 object detectors trained on an own dataset containing images from military environment. The achieved accuracy was 91 \% using the Faster-RCNN and 88 \% using the SSD. Unfortunately, the study is not written in English.

In the paper~\cite{Huang2022}, the authors describe the available efficient deep learning models such as MobileNet or SqueezeNet of various versions together with a facial expression recognition experiment performed on RaspberryPi 3B+ and 4B+ devices with an own reduced-size model called CNN-RIS inspired by DenseNet. The implementation is made publicly available. The proposed network achieved the FPS of 1.38 on RaspberryPi 3 and 5.07 on RaspberryPi 4 with the accuracy and energy consumption comparable to other edge models.

The authors of~\cite{10.1145/3616392.3623417} introduce a benchmark of six CNN models (SSDv1, SSDv2, EfficientDetv0, EfficientDetv1, YOLOv5 and MobileLocalizer) pretrained on the COCO dataset and employed on the RaspberryPi 4B embedded computer complemented with the Google Coral accelerator. All models were tested on a dataset of public traffic videos under various weather and lighting conditions. The examined criteria were set as F1-score, Frames per Second (FPS) and Energy consumption. The highest FPS was achieved using the MobileLocalizer network followed by both SSD models. On the other hand, the highest average F1-score was achieved using the SSDv2 model followed by the YOLOv5. The energy consumption using the Coral accelerator was relatively balanced across the models, with the lowest one achieved by the YOLOv5 model, which paradoxically had the highest energy consumption while running on a CPU.

A detailed comparison of the YOLOv8 model with various backbones employed on both RaspberryPi 4 and RaspberryPi 5 is presented in~\cite{rPi_YOLO}. The presented tutorial also describes size of the networks, their accuracy in the terms of $mAP[0.5-0.95]$ and the inference times. Tutorial also emphasizes the difference between the S and N variants of the YOLO models, where the S variant is approximately $2.5\times$ faster while reaching a 10\% lower mAP score.



\section{Experimental setup}
\label{se:experimentalSetup}

In this section, we first describe the dataset used, which is followed by a description of the experimental description, implementation, evaluation metrics, and experimental hardware.

\subsection{Used dataset}

For the fire sensor detection experiments, we collected more than 2700 photos from various public environments such as public corridors, schools, garages, or shops. As the data was relatively difficult to collect in a sufficient amount for training, it was supplemented by a set of photos captured in a controlled environment, where the fire sensors were placed on several backgrounds and captured from a variety of distances and angles. To avoid potential generalization issues, we used a variety of cameras and lenses.

\begin{figure*}[ht!]
     \centering
     \begin{subfigure}{0.45\textwidth}
         \centering
         \includegraphics[width=\textwidth]{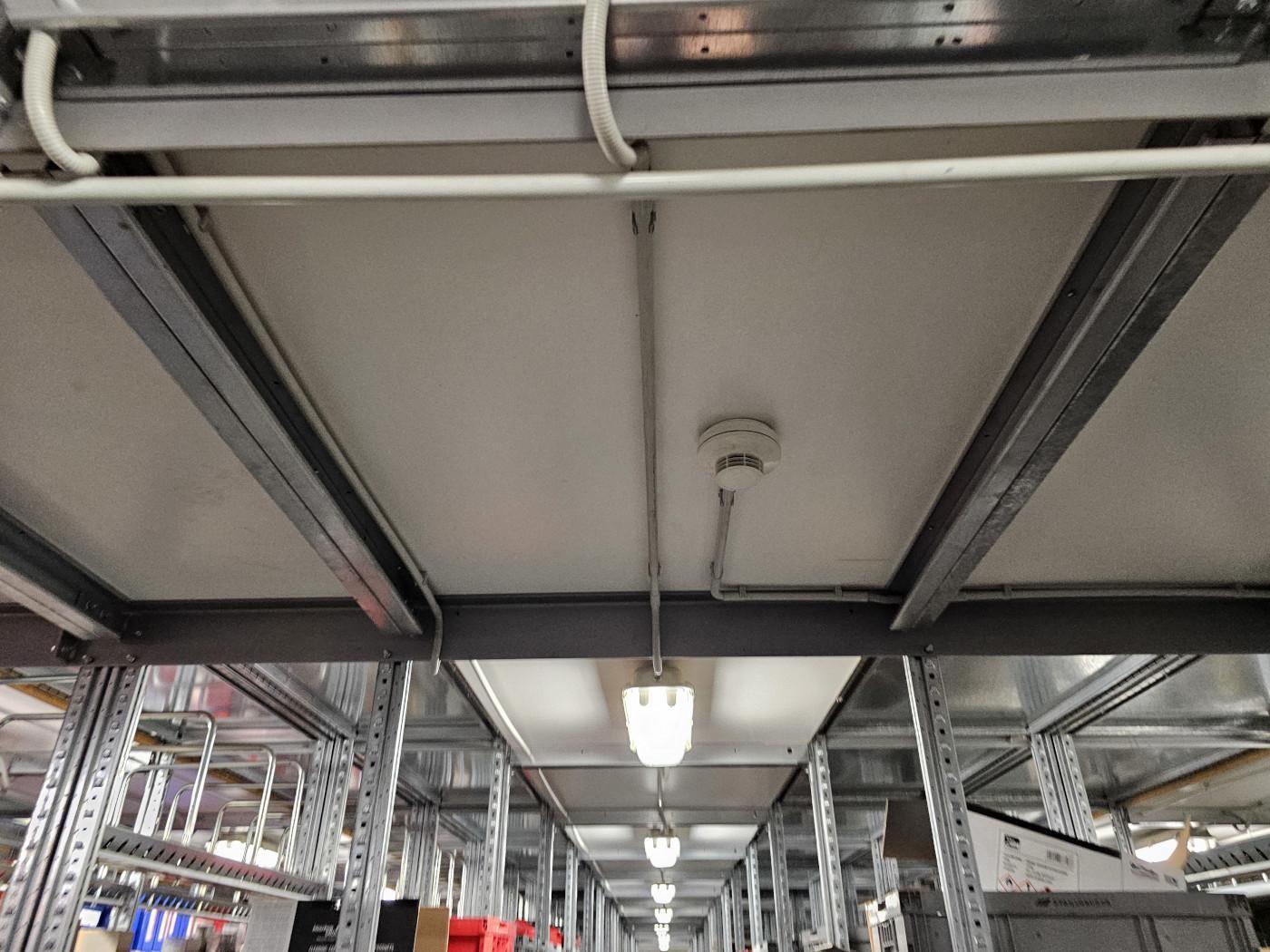}
         \caption{Real sample from the real-world environment.}
         \label{fig:datasetR1}
     \end{subfigure}
     \hfill
     \begin{subfigure}{0.45\textwidth}
         \centering
         \includegraphics[width=\textwidth]{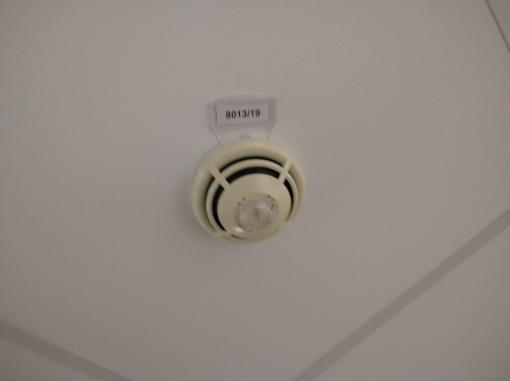}
         \caption{Real sample from the laboratory environment.}
         \label{fig:datasetR2}
     \end{subfigure}
    
     
     \begin{subfigure}{0.45\textwidth}
         \centering
         \includegraphics[width=\textwidth]{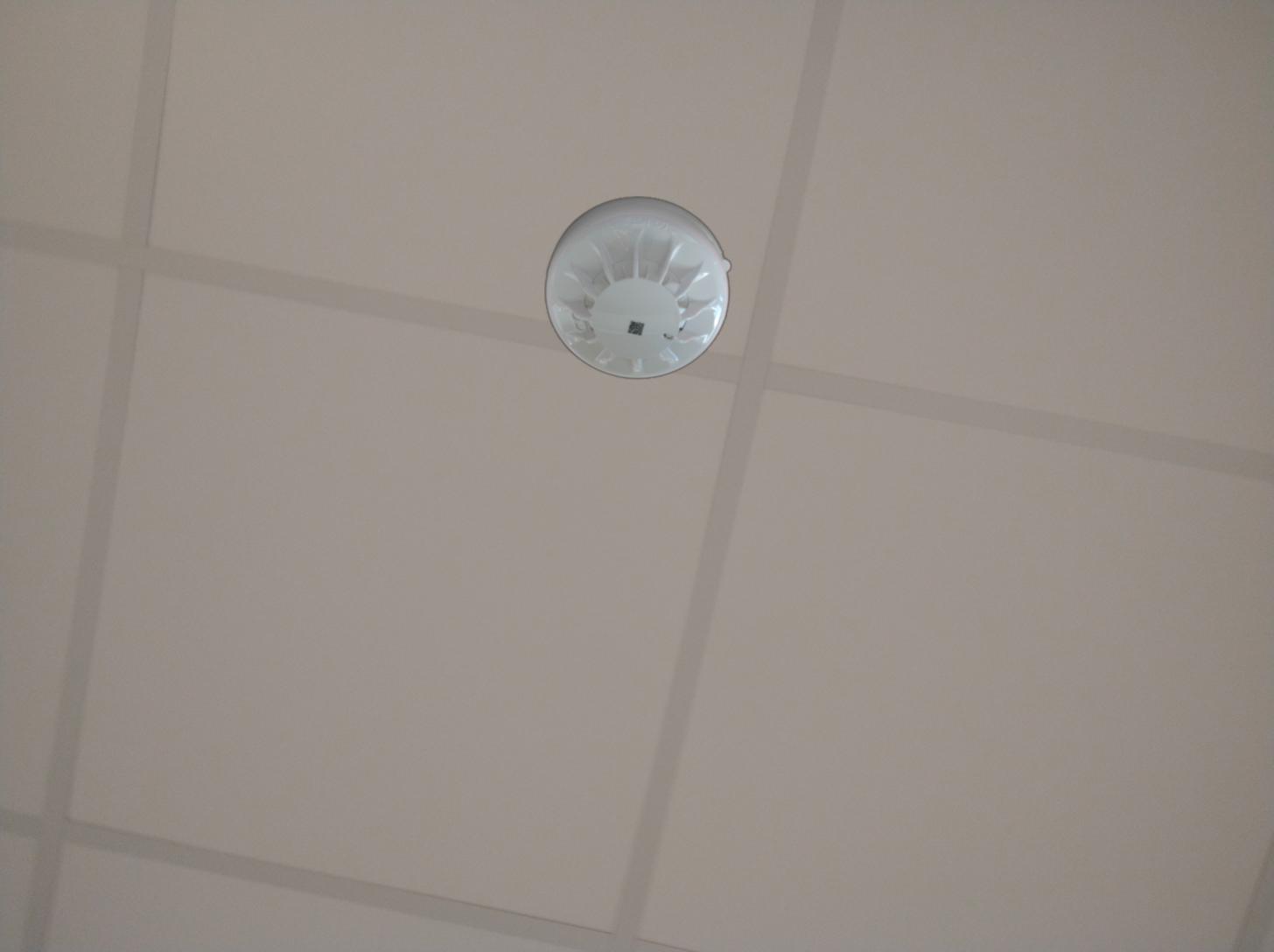}
         \caption{Generated image from real sample.}
         \label{fig:datasetG1}
     \end{subfigure}
     \hfill
     \begin{subfigure}{0.45\textwidth}
         \centering
         \includegraphics[width=\textwidth]{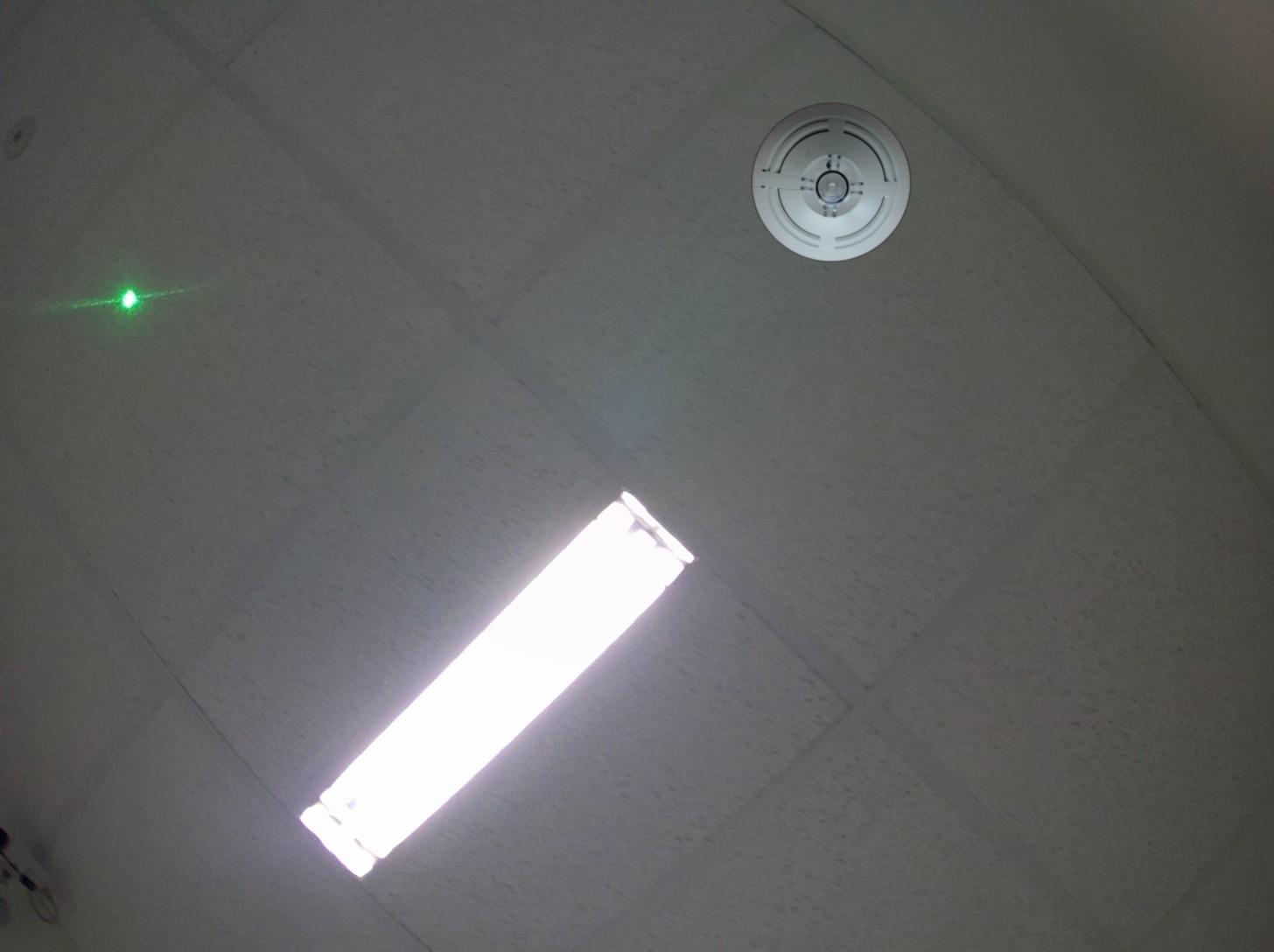}
         \caption{Generated image from rendered sample.}
         \label{fig:datasetG2}
     \end{subfigure}
    
     
     \begin{subfigure}{0.45\textwidth}
         \centering
         \includegraphics[width=\textwidth]{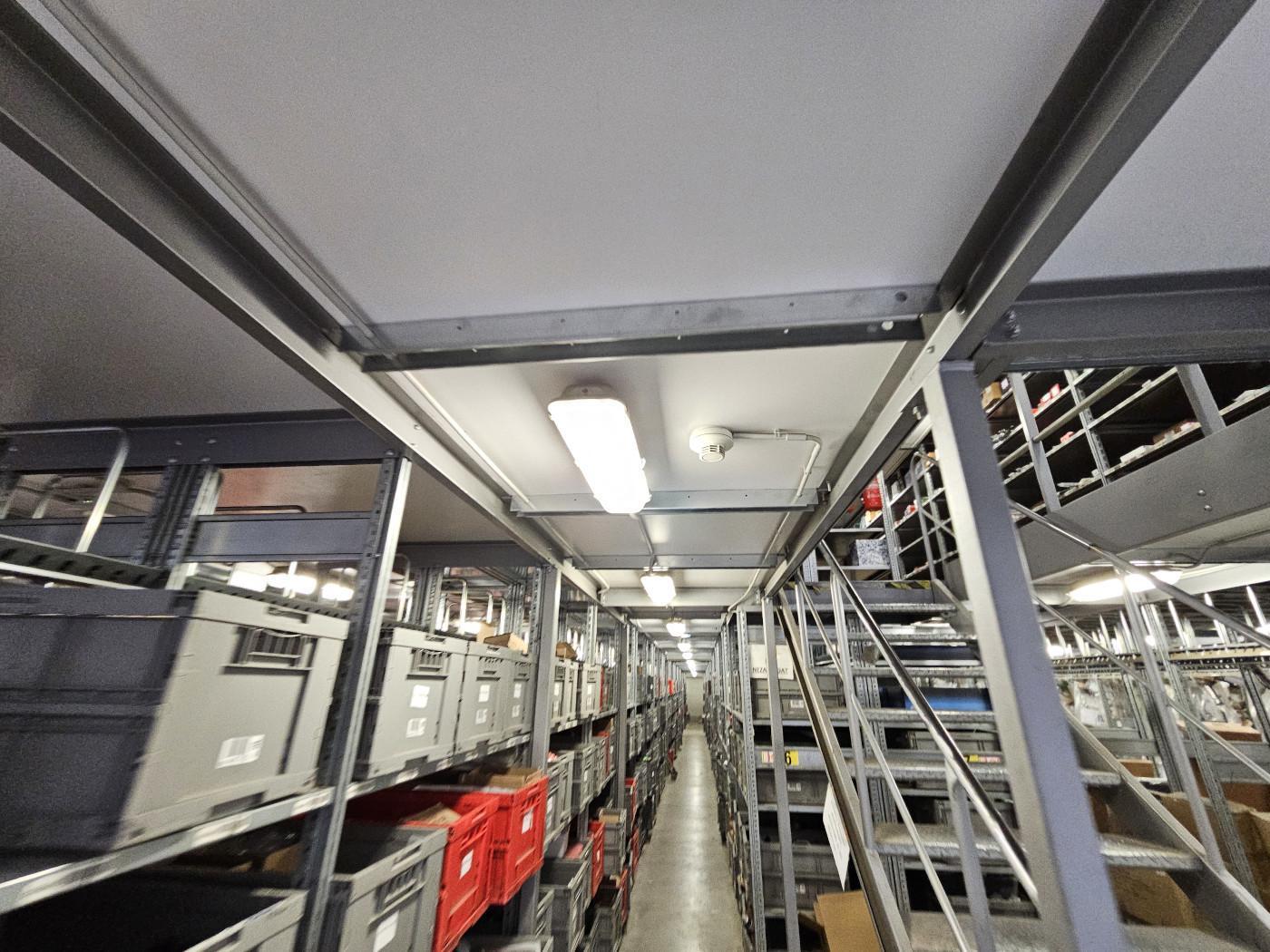}
         \caption{Image from the normal test dataset.}
         \label{fig:datasetTrN}
     \end{subfigure}
     \hfill
     \begin{subfigure}{0.45\textwidth}
         \centering
         \includegraphics[width=\textwidth]{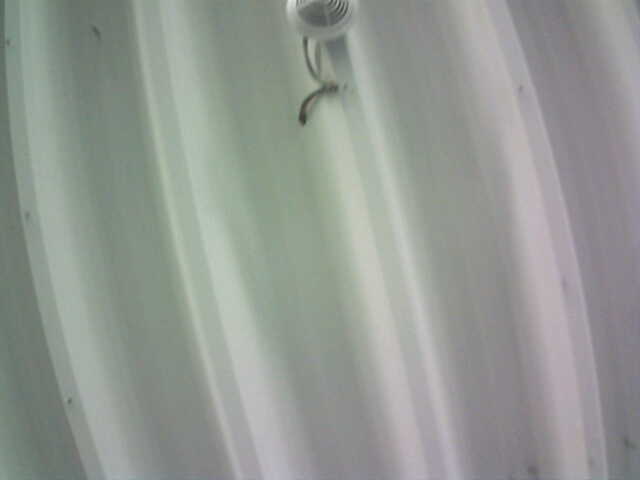}
         \caption{Image from the difficult test dataset.}
         \label{fig:datasetTrD}
     \end{subfigure}
    
    
    \label{fig:datasetR}
\end{figure*}

All captured photos were checked and sorted - photos from the real environment which contained only the background were separated for later use, as were the photos which were blurred, or did not contain the whole sensor. Photos that capture other objects or duplicates were removed from the dataset. In the end, 1672 suitable photos were available. All these photos were manually annotated in the YOLO format and their selection is shown in Fig.~\ref{fig:datasetR1} and Fig.~\ref{fig:datasetR2}.

In order to increase the number of training samples, we decided to use photos or renders of the fire sensors in combination with the real and empty backgrounds. For this purpose, we collected 30 photos of real fire sensors captured from several angles on neutral background and 384 rendered views of two models. These photos were combined with the collection of the real backgrounds mentioned in the previous paragraph in order to create semi-real images. To ensure a more realistic appearance of such images, they were inserted to the upper 40~\% of the background's height with an adaptive brightness adjustment and insertion of a small portion of Gaussian blur to decrease the gradient between the background and the inserted sensor. The chance of inserting two sensors on one background was set as 2~\% and the annotations were generated automatically. In the end, we created an additional 3840 images that were used for training and validation. The generated images are shown in Fig.~\ref{fig:datasetG1} and Fig.~\ref{fig:datasetG2}.


\begin{table}[ht!]
    \caption{Image counts and portions of real and generated data in the experimental datasets.}
	\begin{tabular}{lcccc}
        \toprule
        Dataset & Real-Normal & Real-Difficult & Gen-Real & Gen-Render \\
        \midrule
        Train (\textbf{R}+\textbf{G})  & 1004 & -   & 1536 & 1536  \\ 
        Valid (\textbf{R}+\textbf{G})  & 334  & -   & 384  & 384   \\ 
        Test  (\textbf{R}+\textbf{R})  & 334  & 127 & -    & -     \\ 
		\label{tab:datasetC}
	\end{tabular}
\end{table}

To avoid potential data leak problems and keep the datasets consistent throughout all experiments, we manually create training, validation, and test datasets from the available data as shown in Table~\ref{tab:datasetC}. The split of the real images was set as 60~:~20~:~20 (training~:~validation~:~test) and the generated images 80~:~20 (training~:~validation). In the case of the training and validation datasets, we created one pair of sets from the \textbf{R}eal data and the second pair from the \textbf{G}enerated data which were used throughout the experiments. For a more detailed analysis of the trained models, two test (Test-\textbf{N}ormal and Test-\textbf{D}ifficult) datasets were collected from the available data. The Test-\textbf{N} dataset contains images with standard appearance as used for training and validation, while the Test-\textbf{D} dataset contains of images with difficult appearance such as blur, not complete sensors, or other image distortions. Both testing datasets were used through all performed experiments with their samples shown in Fig.~\ref{fig:datasetTrD}. Our dataset is publicly available online at~\cite{datasetTASDI}.




\subsection{Experiment description}

The aim of this experiment was to evaluate the effect of semi-real images for training and validation. Therefore, we compared the effect of various training sets on detection performance in two test datasets. The training setups were set as follows:

\begin{itemize}
  \item \textbf{Real training set together with real validation set (R-R)}: In this setup, we used real images for both training and validation. No generated images of any kind were used.
  
  \item \textbf{Real training set together with generated validation set (R-G)}: In this setup, we replaced the validation set with a generated one. These images were used from both the generated-real subsets and the generated-render subsets.
  
  \item \textbf{Generated training set together with generated validation set (G-G)}: In this setup, we used both generated training and validation sets. These images were used from both the generated-real subsets and the generated-render subsets.
  
  \item \textbf{Generated training set together with real validation set (G-R)}: In this setup, we replaced the training set with a generated one. These images were used from both the generated-real subsets and the generated-render subsets.
  
  \item \textbf{Mixed real and generated training set together with real validation set (M-R)}: In this setup, we combined both generated and real images for training, and we used real images for validation.
  
\end{itemize}

\bigskip

Depending on the architecture used, we also compared several variants of the selected models and the augmentation framework. As the first step, we determined the optimal dataset combination as described above for the middle-sized variant of the selected object detector. Using the most optimal dataset combination, we compare the performance of the other object detector variants in the second step. Details for individual models are as follows:

\begin{itemize}

  \item \textbf{YOLOv11}: We compare the performance of the N, S, M, and X YOLOv11 models together with the most optimal dataset. The \textit{Randaugment} framework is used as the default augmentation technique.

  \item \textbf{SSD}: We compare the performance of the VGG16 and MobileNetV3-L backbones together with the most optimal dataset. \textit{TorchVision} augmentations \textit{RandomPhotometricDistort}, \textit{RandomZoomOut}, \textit{RandomIoUCrop}, and \textit{RandomHorizontalFlip} are used as the default augmentation techniques.

  \item \textbf{RT-DETRv2}: We compare the performance of the S, M, and L RT-DETRv2 models together with the most optimal dataset. The \textit{Randaugment} framework is used as the default augmentation technique.
\end{itemize}

Finally, to suppress the effect of different default augmentations of the selected object detectors, performance of the most optimal model is compared with its default augmentations and with the techniques simulating the expected real-world distortions in comparison with the default setup of each detector. For these augmentations, we used the \textit{Albumentation} augmentation framework which was set consistently over all experiments. We decided to use augmentations that simulate possible image distortions and degradations occurring during the platform UAV's flight. Namely, we use the following augmentations: \textit{Autocontrast}, \textit{Illumination}, \textit{MotionBlur}, \textit{Defocus}, \textit{ChromaticAberation}, and \textit{ISONoise}. The probability parameter for all augmentations was set to 0.2.

\subsection{Used implementations and evaluation metrics}

For our experiments, we used the Ultralytics YOLOv11 implementation available at~\cite{yoloV11}, the PyTorch SSD implementation available at~\cite{pytorchSSD} and the RT-DETRv2 implementation available at~\cite{pytorchRTDETR}. Training of all models was performed on a work station equipped with the AMD Ryzen 7 7800X3D (8-core) CPU, and NVIDIA GeForce RTX 4080 SUPER (16~GB VRAM) GPU, and with 32~GB of RAM. Training hyperparameters are shown in Table~\ref{tab:hyperparameters}.

\begin{table}[ht!]
    \caption{Training hyperparameters of the selected models.}
	\begin{tabular}{lccc}
        \toprule
        Hyperparameter & YOLOv11 & RT-DETRv2 & SSD \\
        \midrule
        Image size          & 640x640           & 640x640            & 300x300 \\
        Number of epochs    & 120               & 120                & 120 \\
        Batch size          & 16                & 16                 & 16 \\
        Optimizer           & AdamW              & AdamW              & SGD \\
        Learning rate       & $7\times 10^{-4}$              & $2 \times 10^{-5}$ & $1.5 \times 10^{-3}$ \\
        Weight decay        & $5\times 10^{-4}$ & $2\times 10^{-4}$  & $2\times 10^{-5}$ \\
		\label{tab:hyperparameters}
	\end{tabular}
\end{table}

To evaluate the experiments and find the model with the optimal performance and dataset combination, we used the mAP@0.5 score calculated over the validation and both testing datasets. The selection of models for experiments with model variants and custom augmentations was performed based on the maximum average mAP@0.5 score on both testing datasets complemented by the minimal difference of these scores. Fulfillment of these parameters ensures selection of the model with the highest and the most consistent performance.

To improve readability of the \textit{Experimental Results} section, the experimental results are shown in format (mAP@0.5 scores of average test, Test-\textbf{N} and Test-\textbf{D}) without further description.

\subsection{Experimental hardware}
\label{sec:setupHW}

To evaluate the performance of the object detectors, we decided to use embedded computers from the Raspberry Pi (rPi) family together with two proprietary camera modules. To assess suitability, we compare the popular Raspberry Pi 4B model with the more recent Raspberry Pi 5. Although rPi 5 has more computational power in comparison with rPi 4B, it also has greater requirements for power and heat management. Therefore, rPi 4B may still be a suitable choice for some less demanding applications. The parameters of both rPi embedded computers are shown in Table~\ref{tab:rPiParam}. For the both rPi cmputers, we used the Raspberry Pi OS Bookworm operation system.

\begin{table*}[ht!]
    \caption{Parameters of the evaluated rPi embedded computers}
	\begin{center}
    \begin{minipage}{\textwidth}
		\begin{tabular*}{\textwidth}{@{\extracolsep{\fill}}lccccc@{\extracolsep{\fill}}}
		\toprule
        rPi type        & CPU              & Core       & Frequency {[}MHz{]} & GPU           \\
        \toprule
        Raspberry Pi 4B & Broadcom BCM2711 & Cortex-A72 & 1500                & VideoCore VI  \\
        Raspberry Pi 5  & Broadcom BCM2712 & Cortex-A76 & 2400                & VideoCore VII        
		\end{tabular*}
		\label{tab:rPiParam}
    \end{minipage}
	\end{center}
\end{table*}

\section{Experimental results}
\label{sec:experimentResults}

The experimental results of the first experiment focused on the training strategies using the medium sized models are shown in Table~\ref{tab:expResults1} followed by a comparison of various model variants trained with the best-performing strategy in Table~\ref{tab:expResults2} and finally concluded by an analysis of the effect of selected augmentations in Table~\ref{tab:expResults3}.

\begin{table*}[ht!]
    \caption{Detection results over various datasets.}
	\begin{center}
    \begin{minipage}{\textwidth}
		\begin{tabular*}{\textwidth}{@{\extracolsep{\fill}}llccccc@{\extracolsep{\fill}}}
		\toprule
        Training dataset & Validation dataset & \multicolumn{1}{l}{mAP@0.5 Validation} & \multicolumn{1}{l}{mAP@0.5 Test-\textbf{N}} & \multicolumn{1}{l}{mAP@0.5 Test-\textbf{D}} & \multicolumn{1}{l}{mAP@0.5 Test Avg} & \multicolumn{1}{l}{mAP@0.5 Test Diff}\\
        \toprule
        \multicolumn{7}{c}{\textit{YOLOv11m}}   \\
        \midrule
        Real             & Real         & 0.941          & \textbf{0.934 } & 0.688          & 0.811          & 0.246\\
        Real             & Generated    & 0.932          & 0.864           & 0.331          & 0.597          & 0.533\\
        Generated        & Generated    & \textbf{0.995} & 0.495           & 0.276          & 0.385          & 0.219\\
        Generated        & Real         & 0.658          & 0.578           & 0.372          & 0.475          & 0.206\\
        Mixed            & Real         & 0.918          & 0.924           & \textbf{0.731} & \textbf{0.827} & \textbf{0.193}\\
        \midrule
        \multicolumn{7}{c}{\textit{SSD with MobileNetV3 backbone}}   \\
        \midrule
        Real             & Real          & 0.883          & 0.835          & \textbf{0.664} & \textbf{0.749} & 0.171\\
        Real             & Generated     & 0.818          & 0.857          & 0.618          & 0.737          & 0.239\\
        Generated        & Generated     & \textbf{1.000} & 0.435          & 0.346          & 0.390          & \textbf{0.089}\\
        Generated        & Real          & 0.526          & 0.463          & 0.226          & 0.344          & 0.237\\
        Mixed            & Real          & 0.892          & \textbf{0.860} & 0.598          & 0.729          & 0.262\\
        \midrule
        \multicolumn{7}{c}{\textit{RT-DETRv2-M}}   \\
        \midrule
        Real             & Real          & 0.897          & 0.886          & 0.639          & 0.762          & 0.247\\
        Real             & Generated     & 0.703          & 0.751          & 0.525          & 0.638          & 0.226\\
        Generated        & Generated     & \textbf{0.989} & 0.434          & 0.007          & 0.220          & 0.427\\
        Generated        & Real          & 0.667          & 0.643          & 0.156          & 0.399          & 0.487\\
        Mixed            & Real          & 0.917          & \textbf{0.911} & \textbf{0.696} & \textbf{0.803} & \textbf{0.215}
        \end{tabular*}
		\label{tab:expResults1}
    \end{minipage}
	\end{center}
\end{table*}

\begin{table*}[ht!]
    \caption{Detection results over various variants of object detectors.}
	\begin{center}
    \begin{minipage}{\textwidth}
		\begin{tabular*}{\textwidth}{@{\extracolsep{\fill}}llcccc@{\extracolsep{\fill}}}
		\toprule
        Model variant & mAP@0.5 Validation & mAP@0.5 Test-\textbf{N} & mAP@0.5 Test-\textbf{D} & mAP@0.5 Test Avg & mAP@0.5 Test Diff\\
        \toprule
        \multicolumn{6}{c}{\textit{YOLOv11 over the mixed training and real validation dataset.}}   \\
        \midrule
        YOLOv11n        & 0.942          & \textbf{0.940} & \textbf{0.828} & \textbf{0.884} & \textbf{0.112}\\
        YOLOv11s        & \textbf{0.945} & 0.934          & 0.776          & 0.855          & 0.158\\
        YOLOv11m        & 0.918          & 0.924          & 0.744          & 0.834          & 0.180\\ 
        YOLOv11x        & 0.943          & 0.933          & 0.803          & 0.868          & 0.130\\   
        \midrule
        \multicolumn{6}{c}{\textit{SSD over the real training and validation datasets.}}   \\
        \midrule
        MobileNetV3-L   & 0.883          & 0.835          & 0.664          & 0.749          & 0.171\\
        VGG16           & \textbf{0.914} & \textbf{0.909} & \textbf{0.843} & \textbf{0.876} & \textbf{0.066}\\  
        \midrule
        \multicolumn{6}{c}{\textit{RT-DETRv2 over the mixed training and real validation dataset.}}   \\
        \midrule
        RT-DETRv2-S     & 0.870          & 0.847          & 0.531          & 0.689          & 0.316\\
        RT-DETRv2-M     & \textbf{0.917} & \textbf{0.911} & 0.696          & 0.803          & 0.215\\
        RT-DETRv2-L     & 0.876          & 0.897          & \textbf{0.769} & \textbf{0.833} & \textbf{0.128}\\   
        \end{tabular*}
		\label{tab:expResults2}
    \end{minipage}
	\end{center}
\end{table*}

\begin{table}[ht!]
    \caption{Effect of augmentations on evaluated object detectors.}
	\begin{tabular}{lccc}
		\toprule
        Augmentations  & mAP@0.5 Test-\textbf{N}  & mAP@0.5 Test-\textbf{D}  & Test Avg \\
        \toprule
        \multicolumn{4}{c}{\textit{YOLOv11 over the mixed training and real validation dataset.}}   \\
        \midrule
        Default        & \textbf{0.940} & \textbf{0.828} & \textbf{0.884}\\
        Albumentations & 0.936          & 0.815          & 0.875\\
        \midrule
        \multicolumn{4}{c}{\textit{SSD over real training and validation dataset.}}   \\
        \midrule
        Default        & 0.909 & \textbf{0.843} & \textbf{0.876} \\
        Albumentations & \textbf{0.916}          & 0.671          & 0.793 \\
        \midrule
        \multicolumn{4}{c}{\textit{RT-DETRv2 over the mixed training and real validation dataset.}}   \\
        \midrule
        Default        & \textbf{0.897} & \textbf{0.769} & \textbf{0.833} \\
        Albumentations & 0.879          & 0.709          & 0.794 \\
		\label{tab:expResults3}
	\end{tabular}
\end{table}

\subsection{YOLOv11 related experiments}


As the first part of our experiments, we compared the performance of the \textit{YOLOv11m} model over various combinations of input datasets with the aim of discovering the most optimal combination of training and validation. Although the best mAP@0.5 validation score was achieved using the combination of \textbf{G-G} datasets (mAP@0.5 of 0.995), the results of both testing datasets failed (mAP@0.5 scores of 0.385, 0.495 and 0.276), indicating a successful training, but a loss of generalization abilities. The best results were achieved with the \textbf{M-R} experiment (mAP@0.5 scores of 0.827, 0.924 and 0.731) followed by slightly worse results achieved with \textbf{R-R} experiment (mAP@0.5 scores of 0.811, 0.934 and 0.688). These results seem to indicate that the influence of the generated datasets is not as high as expected, but that they might lead to slight improvements if used in combination with real ones.


The second part of the YOLOv11 experiment involved the comparison of various model variants using the \textbf{M-R} combination of datasets. Surprisingly, the best results in the test datasets were achieved with the small variants YOLOv11n (mAP@0.5 scores of 0.884, 0.940 and 0.828) and YOLOv11s (mAP@0.5 scores of 0.855, 0.934 and 0.776), which should be the most suitable for use on an embedded device due to their low computational demands. The YOLOv11n model also achieved the most balanced results with the difference mAP@0.5 in both test datasets of 0.112.


In the third part of our experiment we compared the results achieved with the original augmentations and the \textit{Albumentations} framework. The default setup achieved slightly better, but almost comparable results (mAP@0.5 scores of 0.884, 0.940 and 0.828 using the default setup in comparison with mAP@0.5 scores of 0.875, 0.936 and 0.815 using the Albumentations).

\subsection{SSD related experiments}


In the first part, we compared the performance of the SSD object detector with the MobileNetV3-L backbone over various combinations of training and validation datasets. The best validation scores were achieved in the \textbf{G-G} experiment with the resulting mAP@0.5 score of 1.00 and the test results (mAP@0.5 scores of 0.390, 0.435 and 0.346) again indicating a successful training with a loss of generalization as in the case of YOLOv11. The best results over the test datasets were achieved with the \textbf{R-R} experiment (mAP@0.5 scores of 0.749, 0.835 and 0.664) followed by the \textbf{M-R} experiment (mAP@0.5 scores of 0.729, 0.860 and 0.598). The smallest difference in test mAP@0.5 was achieved in the \textbf{G-G} experiment, but the result is not relevant due to the small test scores achieved there.


For the second part, we compared two SSD backbones on the \textbf{R-R} dataset combination. In this case, the VGG16 backbone outperformed MobileNetV3-L in all validation and test scores, especially in the difficult test dataset (mAP@0.5 scores of 0.876, 0.909 and 0.843), also achieving a very small difference in test mAP@0.5 of 0.066.


In the third part of our experiment we compared the results achieved with the original augmentations and the \textit{Albumentations} framework. In this case, the use of the \textit{Albumentation} library led to slightly better results on the normal test dataset compared to the default setting, but also into the failure in the difficult test dataset on the other hand (mAP@0.5 scores of 0.916, 0.671 and 0.793).

\subsection{RT-DETRv2 related experiments}


As in the previous cases, the best results over the validation dataset were achieved using the \textbf{G-G} experiment with the resulting mAP@0.5 of 0.989. Nevertheless, this combination performed as the worst on both testing datasets (mAP@0.5 scores of 0.220, 0.434 and 0.007) while the best test results were achieved in the \textbf{M-R} experiment (mAP@0.5 scores of 0.803, 0.911 and 0.696). The results indicate again a loss of generalization while using the generated datasets for both training and validation, but they show a possible positive effect of the generated images being used for training in combination with the real ones.


In the second part of our experiment, we compared the performance of the RT-DETRv2 variants on the \textbf{M-R} dataset combination. The best results on the validation dataset were achieved with the medium RT-DETRv2-M using the ResNet34 backbone with the resulting mAP@0.5 scores of 0.917, while the best test results achieved the RT-DETRv2-L using the ResNet50 backbone (mAP@0.5 scores of 0.833, 0.897 and 0.769).


In the third part of our experiment we compared the results achieved with the original augmentations and the \textit{Albumentations} framework. The results show that custom augmentations achieved slightly worse results compared to the default setup (mAP@0.5 scores of 0.794, 0.879 and 0.709).

\section{Discussion}
\label{sec:discussion}

\subsection{Comparison of training approaches}

As a first research question, we aim to discuss the effect of synthetic data and the training strategy on the achieved results. In general, adding the synthetic data has effect only in combination with the real data. From this point of view, this process could be considered as a kind of data augmentation. Evaluating the experimental results, we assume that the combination of the real parts of the training and validation dataset was sufficient for a successful training of all detectors, as the \textbf{R-R} experiments achieved sufficient mAP@0.5 scores. The \textbf{R-R} experiment brought the best results in the case of the SSD object detector and the second best results in the cases of the YOLOv11 and RT-DETRv2 detectors.

In the case of the SSD object detector and the \textbf{R-G} experiment, synthetic data also performed relatively well as the validation dataset, achieving the second best result. On the other hand, this experiment produced comparable results for the YOLOv11 and RT-DETRv2 detectors in the case of the validation and normal test dataset results, but significantly worse results in the case of the difficult test dataset. Based on the results achieved, we assume that using the synthetic data for validation is not as robust to handle non-standard looking images as the generated data used for validation are much more simple.

Using of the generated data for training and validation (\textbf{G-G} experiments) leads to overfitting and failure on both testing datasets. The incorporation of a limited number of renders and studio photos of the sensors into the real backgrounds therefore did not bring about a sufficient amount of information needed for successful training of the examined detectors. As the renders were generated in the same way and the studio photos were captured by the same camera, a domain shift might also worsen the achieved results.

When the generated data are combined with the real ones for training (\textbf{M-R} experiment), we see an improvement in the results achieved. This improvement compared to the \textbf{R-R} combination is the most prominent in the case of the YOLOv11 and RT-DETRv2 results. Considering RT-DETRv2, the use of the generated data for training together with the real data leads to a most significant improvement compared to the convolutional-based detectors. As the transformer backbones prioritize the global context and are more sensitive to the amount of input data, increasing the training dataset by combining real and generated data seems to be beneficial. It is also highly probable that transformer-based architectures will benefit more from the generated data compared to convolution-based detectors but is more demanding to the amount of training data.

\subsection{Comparison of the tested models}

As the second research question, we aim to compare evaluated object detection models together with their different variants. The problem is specific as we detect only one object class while the used detectors are designed to recognize a high variety of possible classes.

\textbf{YOLOv11 related experiments:} All variants of the YOLO detector performed similarly on the validation dataset achieving mAP@0.5 score of over 0.91 with the best mAP@0.5 score of 0.945 achieved with the YOLOv11s. The results achieved on the normal test dataset also did not differ significantly, while the best mAP@0.5 score of 0.940 was achieved with YOLOv11n. The highest variance was achieved on the difficult test dataset with the best mAP@0.5 score of 0.828 on the YOLOv11n.

Based on the results achieved, we assume that the larger YOLO variants are overfitted or that they require more training data. Smaller N and S models work better with the limited available dataset and under blur and sub-optimal illumination. This is relevant as drones often face vibration and these kinds of conditions.

\textbf{SSD related experiments:} In the case of the SSD object detector, we compared the MobileNetV3 and VGG16 backbones. The VGG16 backbone was superior to the MobileNetV3 all datasets. The mAP@0.5 score achieved on the validation dataset was 0.883 for MobileNetV3 and 0.914 for VGG16. The best mAP@0.5 scores achieved in the testing datasets were achieved with the VGG16 backbone: 0.909 in the case of the normal one and 0.843 in the case of the difficult one.

In the case of the SSD, backbone size is a crucial factor in order to reach more reliable detections. On the other hand, using the VGG16 will result in significantly higher inference times in comparison with MobileNetV3.

\textbf{RT-DETRv2 related experiments:} Although the best results on the validation and normal test dataset were achieved with the M variant of the model (mAP@0.5 score of 0.917 for validation and 0.911 for normal test dataset), the L variant of the model performed better on the difficult test dataset achieving the mAP@0.5 score of 0.769. Together with a good result on the normal test dataset, it also resulted in a higher average mAP@0.5 score over both test dataset, where the L model proved to be superior over the M one (average mAP@0.5 score of 0.833 in comparison with 0.803).

The transformer based RT-DETRv2 object detector is more demanding on the size of the training dataset as there is also a largest gap between the results of the \textbf{R-R} and \textbf{M-R} experiment. We interpret the better results over the validation and normal test datasets achieved with the M variant such that the dataset size is not completely sufficient to train the L model variant. On the other hand, the L variant is more robust in complex real-world conditions as the results in the difficult dataset are good enough even to increase the average mAP@0.5 test score over the results achieved with the M model variant.

\subsection{Effect of the \textit{Albumentation} framework}

As the third research question, we discuss the effect of uniform augmentations over the best performing object detectors from the previous experiment. The custom augmentations using the Albumentations framework outperformed the default setup only in the case of the SSD detector with the VGG16 backbone on the normal test dataset.

The smallest difference in the resulting mAP@0.5 scores was achieved with YOLOv11 that achieved the average mAP@0.5 score of 0.875 compared to the mAP@0.5 score of 0.884 with default augmentations. This model was followed with an object detector RT-DETRv2-L with an average mAP@0.5 score of custom augmentations of 0.794 compared to an mAP@0.5 score of 0.833 with the default augmentations. Finally, the SSD model achieved an average mAP@0.5 score of 0.876 compared to the mAP@0.5 score of 0.793 with default augmentations and highest slump in the case of the difficult test dataset showing the smallest potential for using custom augmentations.

In the case of YOLOv11 as the model with the smallest difference of the test mAP@0.5 scores, we performed an additional experiment, where we combined the default augmentations with the Albumentation framework. Surprisingly, the results obtained were worse in comparison with the both default and pure Albumentations experiment (average mAP@0.5 of 0.824, test mAP@0.5 of 0.937 and 0.712) while the largest drop on the difficult test dataset. We explain those results by making the augmentations too complex and by a slight overfit on appearance of the standard looking sensors, which worsens the results on difficult data.

From the achieved results, we assume that the advanced augmentations as motion blur or defocus do not bring the expected effect on any of the tested object detectors as their best outcome was only slightly worse, but relatively still comparable results in the case of the YOLOv11 model. The most significant negative effect of these augmentations to the SSD on difficult test dataset might be caused by too high complexity of the VGG16 backbone and disrupting of the its features by too aggresive augmentations. In the case of the RT-DETRv2 and the combined YOLOv11 experiment, the worsening of the results is more significant in the case of the difficult test dataset, which indicates a possible overfit of both models on data with standard appearance and some loss of generalization abilities.

\subsection{Integration to the drone detection system and inference times}
\label{sec:integration}

\begin{figure*}[ht!]
	\centering
	\includegraphics[width=1.0\linewidth]{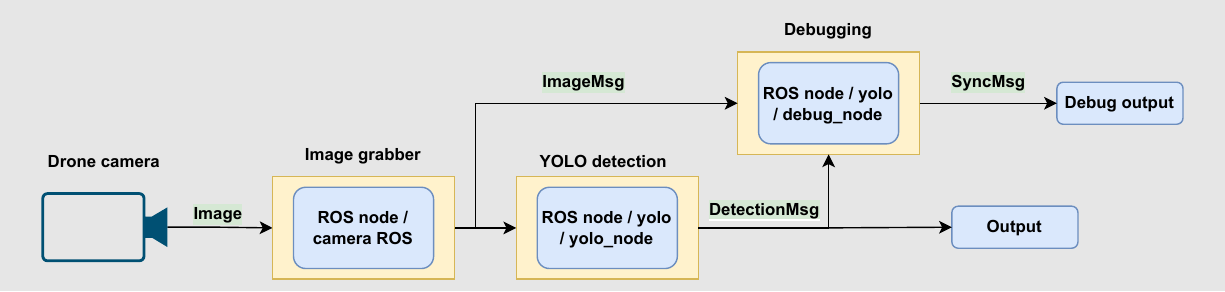}
	\caption{Structure of the proposed ROS-2 pipeline.}
	\label{fig:SchROS}
\end{figure*}

For a future integration into the drone detection system, we implemented a Robot Operating System 2 (ROS-2) pipeline with the structure shown in Fig.~\ref{fig:SchROS}. It consists of the three nodes, each of them responsible for a different task. The first is named \textit{camera\_ros} and it grabs images from camera to image message. The second is named \textit{yolo\_node} and provides the interface of the YOLO model. The last, named \textit{debug\_node}, is intended for debugging purposes and synchronizes detection with the images. The \textit{yolo\_node} wraps the YOLOv11n model in this moment, but could easily be replaced with another variant of the YOLOv11 or with another object detector. Our code is publicly available at~\cite{gitTASDI}.

To evaluate the inference times of the most optimal detectors, we evaluated the performance of both rPi systems described in Section~\ref{sec:setupHW}. The evaluation was performed over the entire ROS-2 pipeline, excluding the debugging node, and it is summarized in Table~\ref{tab:rPiSpeed}. From the perspective of the evaluated HW, the rPi 5 outperformed the older model in all three evaluated detectors, reaching the 3.5 times higher average possible FPS rates. The best inference times were reached with the YOLOv11n model, where the processing times on the rPi 5 allow sufficient FPS rates for the suggested application of fine navigation in lower speeds. On the other hand, the SSD and transformer based RT-DETRv2 are not suitable for this application due to high inference times in the order of seconds.

\begin{table*}[ht!]
    \caption{Inference times comparison of \textit{YOLOv11n} object detector on rPi systems using the image of 320x320 px.}
	\begin{center}
    \begin{minipage}{\textwidth}
		\begin{tabular*}{\textwidth}{@{\extracolsep{\fill}}lcccccc@{\extracolsep{\fill}}}
		\toprule
        rPi type        & preprocess {[}ms{]} & interference {[}ms{]} & postprocess {[}ms{]} & total inference time {[}ms{]} & possible FPS rate \\
        \toprule
        \multicolumn{6}{c}{\textit{YOLOv11n}}   \\
        \midrule
        Raspberry Pi 4B & 9.8 & 597.5 & 2.1 & 609.4  & 1.640 \\
        Raspberry Pi 5  & \textbf{3.1} & \textbf{153.5} & \textbf{0.5 }& \textbf{157.1}  & \textbf{6.365} \\
        \midrule
        \multicolumn{6}{c}{\textit{SSD with VGG16 backbone}}   \\
        \midrule
        Raspberry Pi 4B & 40.5 & 4414.9 & 0.8 & 4456.2  & 0.224 \\
        Raspberry Pi 5  & 3.7  & 1175.0 & 0.3 & 1179.0  & 0.848 \\
        \midrule
        \multicolumn{6}{c}{\textit{RT-DETRv2-L}}   \\
        \midrule
        Raspberry Pi 4B & 119.4 & 15 231.3 & 9.4 & 15 360.1 & 0.065 \\
        Raspberry Pi 5  & 35.4  & 3 727.6  & 0.7 & 3 763.7  & 0.265 \\
		\end{tabular*}
		\label{tab:rPiSpeed}
    \end{minipage}
	\end{center}
\end{table*}

\subsection{Further development and limitations of our study}

The most limiting factor of our study is the relatively small training dataset, as collecting a larger one is very challenging in the industrial environment, especially without an automatic detection platform. In the future, using a prototype of the drone detection system, an extensive dataset from the real-world environment could be collected to solve this issue and to allow more complex models such as RT-DETRv2 or YOLOv11x to be trained according to their possibilities. In addition, a more complex way to generate synthetic data with more respect to the shadows and sensor position could be developed using domain adaptation or style transfer techniques. This should be followed by reducing the expected drone motion blur by its simulation and suppression.

\begin{figure}[ht!]
	\centering
	\includegraphics[width=1.0\linewidth]{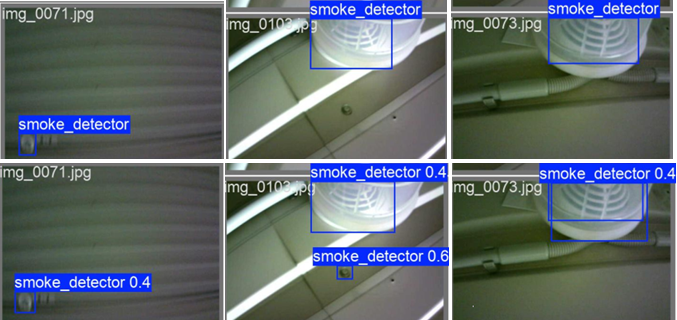}
	\caption{Low confidence YOLOv11n detections.}
	\label{fig:YOLORes}
\end{figure}

An important thing to consider appears after a further analysis of the YOLOv11 results, where we found that most of the low-score detections still appear directly on the sensor with a bounding box of slightly smaller or bigger size as shown in~\ref{fig:YOLORes}. From this analysis, we expect that the real usability of our system might be larger than that indicated by the experimental results. Experimenting with the training process also indicates an effect of the number of epochs, where a higher number of training epochs seems to have an effect on decreasing the FP results on the YOLOv11n model while keeping the same number of TPs. On the other hand, decreasing the number of epochs on the bigger YOLO models seems to improve the results in this case, indicating a possible overfitting while using a large number of model parameters.

Future research should also aim for the integration of our detection framework into the complex drone detection system. Such a system have to include a mechatronic system to perform and evaluate the actual sensor inspection, which is done by a special aerosol and evaluated by the sensor activation. Another important problem is a reliable inner navigation, which will have to be enhanced by other sensors such as lidars to detect possible obstacles, as the navigation system cannot fully rely on a pre-known map with the known obstacles. The use of lidar sensors would ensure a continuous scan of the drone's environment and provide early warning to possible collisions. The drone's trajectory can thus be adjusted in real time to avoid dangerous situations. Assuming stable obstacles within the lidar range, it would also be possible to extend the obstacle detection application with SLAM (Simultaneous Localization and Mapping) navigation. An accurate and reliable position estimation is crucial in environments where radio signal-based localization methods, such as GPS, cannot be used.

A limitation of our study to be considered is that we experiment only with one transformer-based detector. This is because a larger number of these detectors suitable for use in embedded devices are not available due to the training complexity and higher computational demands. Nevertheless, the addition of new detectors might be beneficial in future research.

All datasets were collected in European locations, which might have additional limitations as the appearance of the industrial environment and the smoke sensors used differ worldwide. Generated images also do not respect illumination and the spatial distribution of the background as they might, which could cause another bias to the training process.

Another thing to consider when selecting the model is the licensing management. YOLO models from the YOLOv5 model are licensed under AGPL-3.0 license, which limits the potential commercial use of the frameworks integrating these object detectors. On the other hand, the Pytorch SSD implementation is licensed under the MIT license, and RT-DETRv2 is licensed under Apache-2.0, which allows the integration of these detectors into commercial products without additional requirements. For YOLOv11, a paid commercial license is available.

\section{Conclusion}
\label{sec:conclussion}

In this deployment study, we present a smoke sensor detection system designed to be used on the embedded platform and trained on the limited dataset. We explored possibilities and limitations of five training strategies using both real and synthetic data, compared several variants of the used object detectors, and explored the effect of selected augmentations simulating the expected measurement conditions. The evaluation was performed on two test datasets with a normal and difficult appearance. Besides this, we also present an ROS-2 pipeline, which is ready to be deployed on the rPi system, and we analyze the inference times achieved with this node and the most optimal object detectors. In general, our suggested approach is not limited only to the detection of smoke sensors but also to all practical applications with limited access to training data, which are often difficult to collect on scales of hundreds or thousands of samples.

Although the dataset used is relatively limited, it presents an example of a real-world industrial dataset. In such cases, manual collection of larger amounts of samples is very difficult as it involves work in heights, restricted areas, etc. Therefore, this research presents an example of data-efficient learning, where the existing models are deployed on the limited datasets. Using the synthetic data in the \textbf{M-R} experiment shows a performance improvement of the object detectors used. This effect appeared in the results achieved with the YOLOv11 and RT-DETRv2 models, where it was more significant on the transformer-based RT-DETRv2. Generated data seems to work in a similar manner as augmentations improving the detector's performance by bringing new combinations, views, and scenes together with a higher amount of the training data. On the other hand, it does not bring about so much new information to the training process, which most likely causes failures of the \textbf{G-G} experiments on the test data. If combined with the real data for training and validation to the \textbf{M-R} experiment, the benefits of having more data and training on the real ones merge together.

Considering the average mAP@0.5 scores achieved over all object detectors, the best performing one is the YOLOv11n model trained with the \textbf{M-R} datasets achieving the average mAP@0.5 score of 0.884 and the test difference mAP@0.5 score of 0.112. The second is the SSD model with the VGG16 backbone, which achieves the average mAP@0.5 score of 0.876 and the test difference mAP@0.5 score of 0.066 followed by the RT-DETRv2-L with the average mAP@0.5 score of 0.833 and the test difference mAP@0.5 score of 0.128. Considering the test difference score, the most consistent results between both testing datasets were achieved using the SSD model with the VGG16 backbone. In the case of the best performing YOLOv11, the smaller variants of the model proved to be more robust against overfitting. 

An interesting result is that the large variant of the RT-DETRv2 was found to perform better on the difficult test dataset although the medium variant performed better on both both validation and the normal test dataset. This effect could be explained in such a way that the larger variant is underfitted but still performs more robustly on the difficult and not complete data, as is shown in its smaller mAP@0.5 score difference between both test datasets. A surprising outcome is also that the use of custom augmentations that simulate the expected in-field conditions did not bring the expected improvement but worsened the results through all tested models.

Although the SSD object detector is relatively outdated, the use of the VGG16 backbone brought usable results that outperformed the average performance of the transformer-based RT-DETRv2. As the YOLO family of detectors is not licensed for free commercial use and transformer detectors are more demanding on the size of the input dataset, SSD could still be considered for deployment in the proposed application. On the other hand, results achieved with the custom augmentations indicates potential problems while detecting blurry or defocused images using this detector. The inference achieved with the SSD detector is also comparable to the results of YOLOv11n.

Finally, we introduce an ROS-2 pipeline intended for deployment on an embedded device. Analyzing the inference times, we could explain the FPS rate slightly above 6.3 using the Raspberry Pi 5 embedded computer with the YOLOv11n object detector, which is sufficient for an expected use scenario of fine navigation in low speeds. In the need of faster inference, an HW accelerator or a more computationally efficient device such as the NVIDIA Jetson family could be used. Based on the results achieved, transformer-based detectors are not suitable for deployment in embedded devices because of high computational demands and inference times.


%



\section*{Acknowledgment}

The authors thank all our families, friends, and everyone who supported us during the writing and submission process.

The completion of this paper was made possible by the grant No. FEKT-S-26-8988 - "Advanced Methods in Cybernetics, Robotics, and Artificial Intelligence" financially supported by the Internal science fund of Brno University of Technology.

\bigskip
\bigskip
\bigskip

\ifCLASSOPTIONcaptionsoff
  \newpage
\fi



%




\bibliographystyle{IEEEtran} 


%

\begin{IEEEbiography}[{\includegraphics[width=1in,height=1.25in,clip,keepaspectratio]{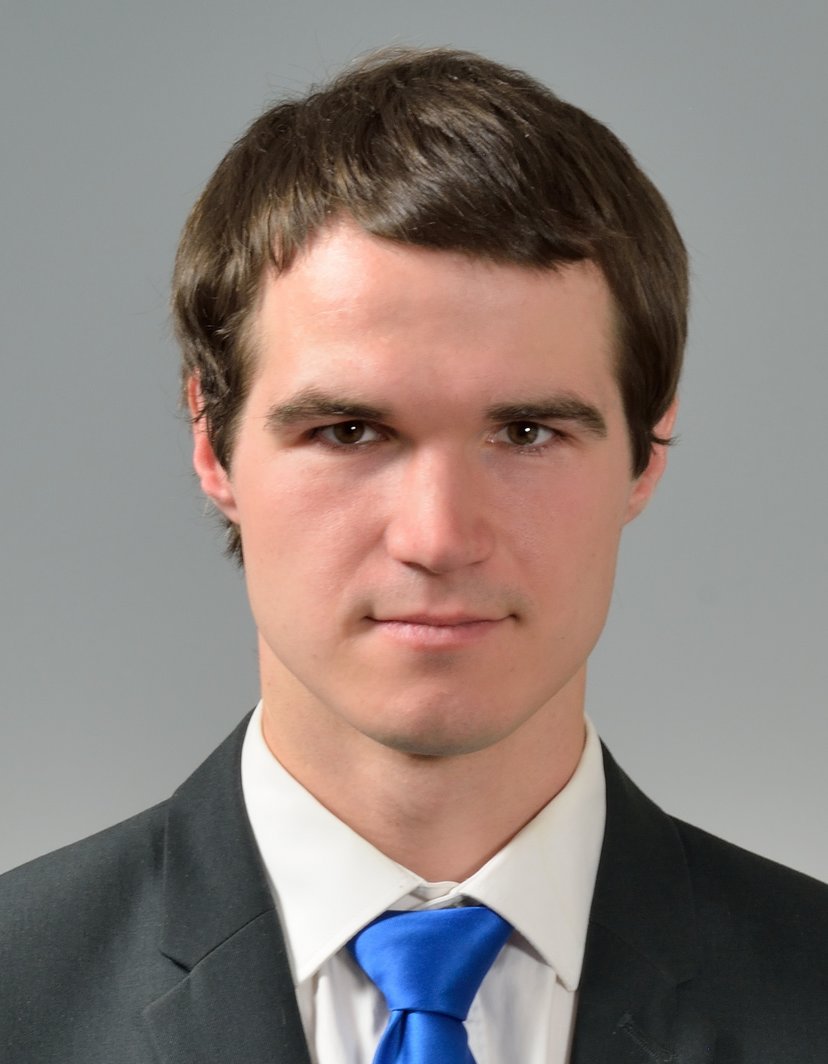}}]{Lukas Kratochvila}
Lukas Kratochvila is a PhD student in the Machine Vision Group of the Department of Control and Instrumentation, FEEC, Brno University of Technology. He graduated with his Master's within the same group in 2019. The aim of his research is visual object detection, motion tracking and the hardware for the AI tasks. He took part in the Erasmus + program and gained countless experiences at the Norwegian University of Science and Technology in Trondheim; for example, in the image style transfer using the CNNs.
\end{IEEEbiography}

\begin{IEEEbiography}[{\includegraphics[width=1in,height=1.25in,clip,keepaspectratio]{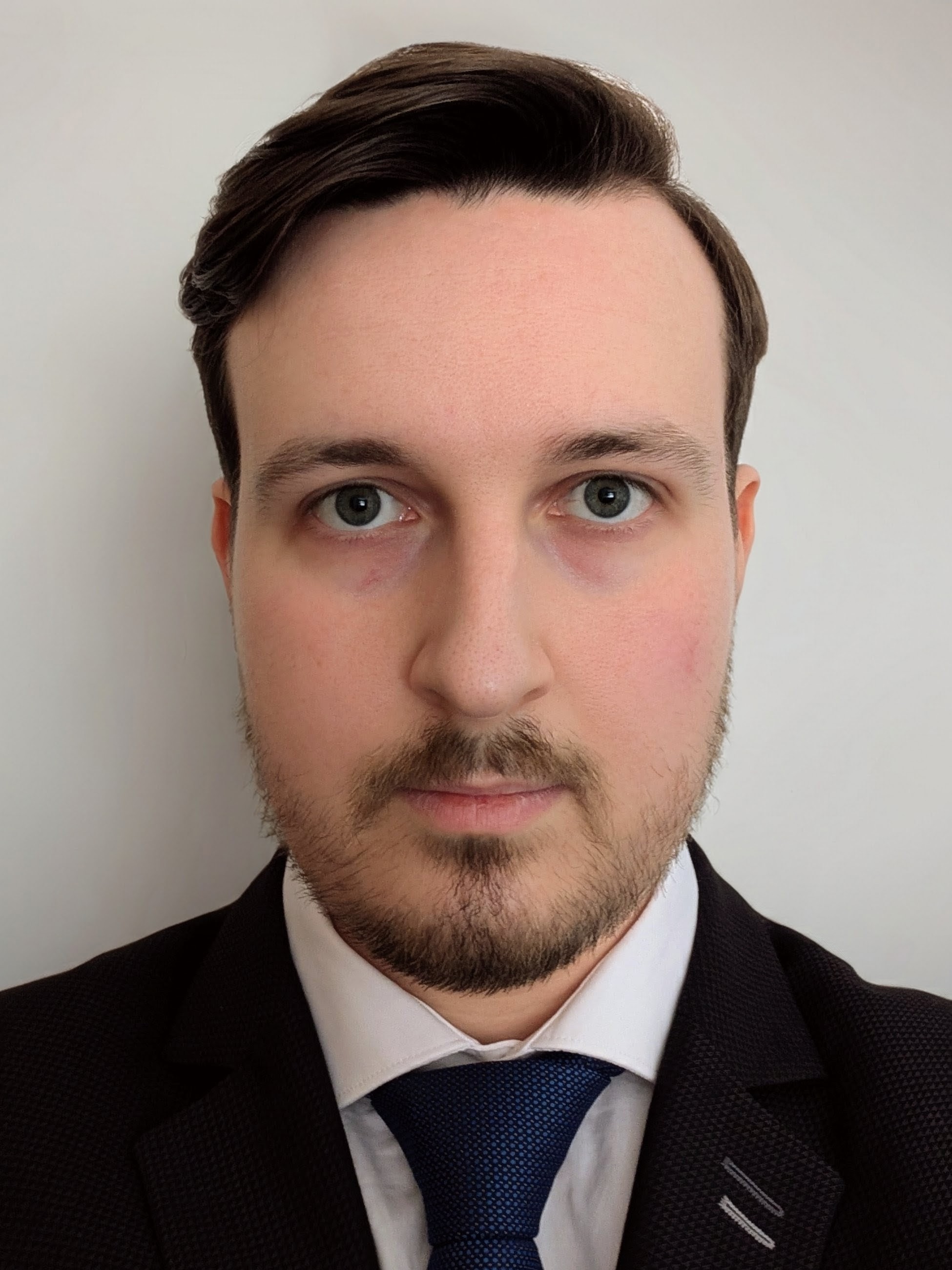}}]{Jakub Stefansky}
Jakub Stefansky is a PhD student in the Signal Lab team of the Department of cybernetics and biomedical engineering, FEECS, Vysoka skola banska - Technical university of Ostrava. He graduated with his master’s degree from the same department in 2021. His research focuses on visual defect detection, semi-supervised learning for biological datasets, medical image processing, and MRI segmentation. He participated in the Erasmus+ program and gained experience in detecting parasites in phytoplankton using semi-supervised learning methods at Lappeenranta-Lahti University of Technology in Finland.
\end{IEEEbiography}

\begin{IEEEbiography}[{\includegraphics[width=1in,height=1.25in,clip,keepaspectratio]{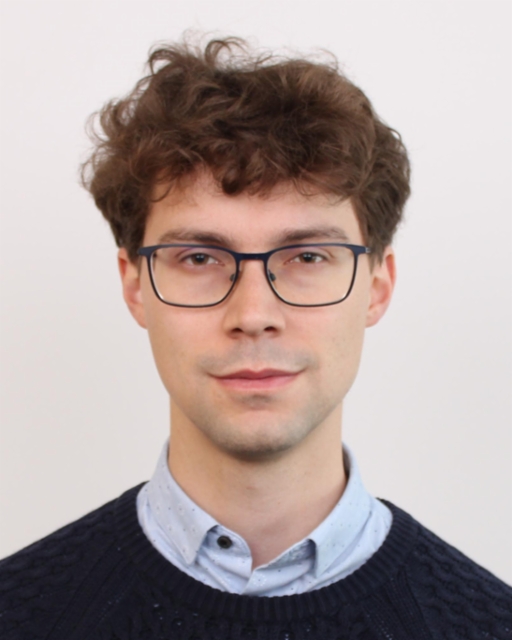}}]{Simon Bilik}
Simon Bilik received the PhD degree in technical cybernetics and computer vision from the Department of Control and Instrumentation, Brno University of Technology, Czech republic and the Computer Vision and Pattern Recognition Laboratory, LUT University, Finland, in 2024. His research field includes applied machine vision and machine learning. Currently, he works as researcher at the Institute for Research and Applications of Fuzzy Modeling of the University of Ostrava together with the research position at Department of Informatics of the Mendel University in Brno.
\end{IEEEbiography}


\begin{IEEEbiography}[{\includegraphics[width=1in,height=1.25in,clip,keepaspectratio]{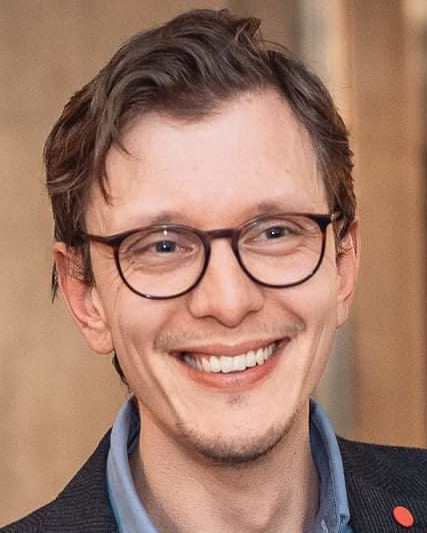}}]{Robert Rous}
Robert Rous is a PhD researcher at the Department of Informatics of Mendel University in Brno, Czech Republic. His research focuses on spectral imaging, computer vision, and AI/ML methods. He is currently engaged in research projects within the Agrivision Research group and contributes to the ESA-backed student project CIMER.
\end{IEEEbiography}

\begin{IEEEbiography}[{\includegraphics[width=1in,height=1.25in,clip,keepaspectratio]{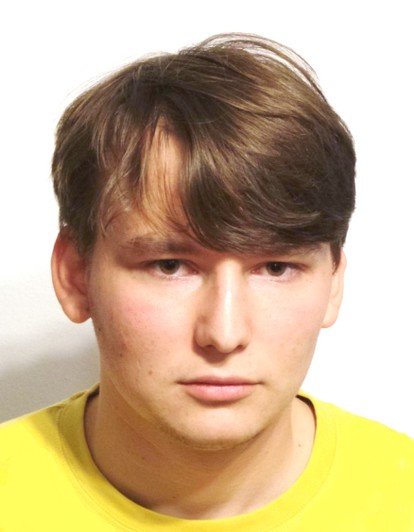}}]{Tomas Zemcik}
Tomas Zemcik is a PhD student in the Machine Vision Group of the Department of Control and Instrumentation, FEEC, Brno University of Technology. He graduated with his Master's within the same group and department in 2019. For five years he has been with CAMEA, spol. s r.o. (a company specialising in image and signal processing in industry and traffic applications) as a developer. His experience also includes involvement in several projects including V3C. During his graduate studies he participated in the Erasmus+ programme and spent a year at the Tampere University of Technology. 
\end{IEEEbiography}

\begin{IEEEbiography}
[{\includegraphics[width=1in,height=1.25in,clip,keepaspectratio]{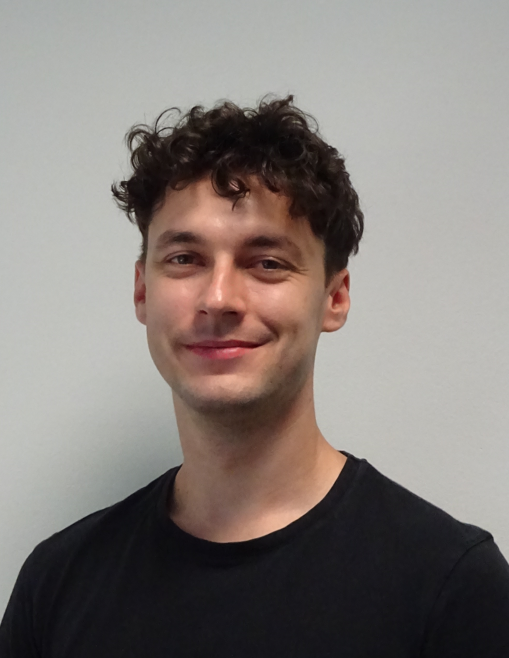}}]{Michal Wolny}
Michal Wolny obtained his MSc degree in Cybernetics, Automation and Measurement, Brno University of Technology. The aim of his research is image processing and object detection.
\end{IEEEbiography}

\begin{IEEEbiography}[{\includegraphics[width=1in,height=1.25in,clip,keepaspectratio]{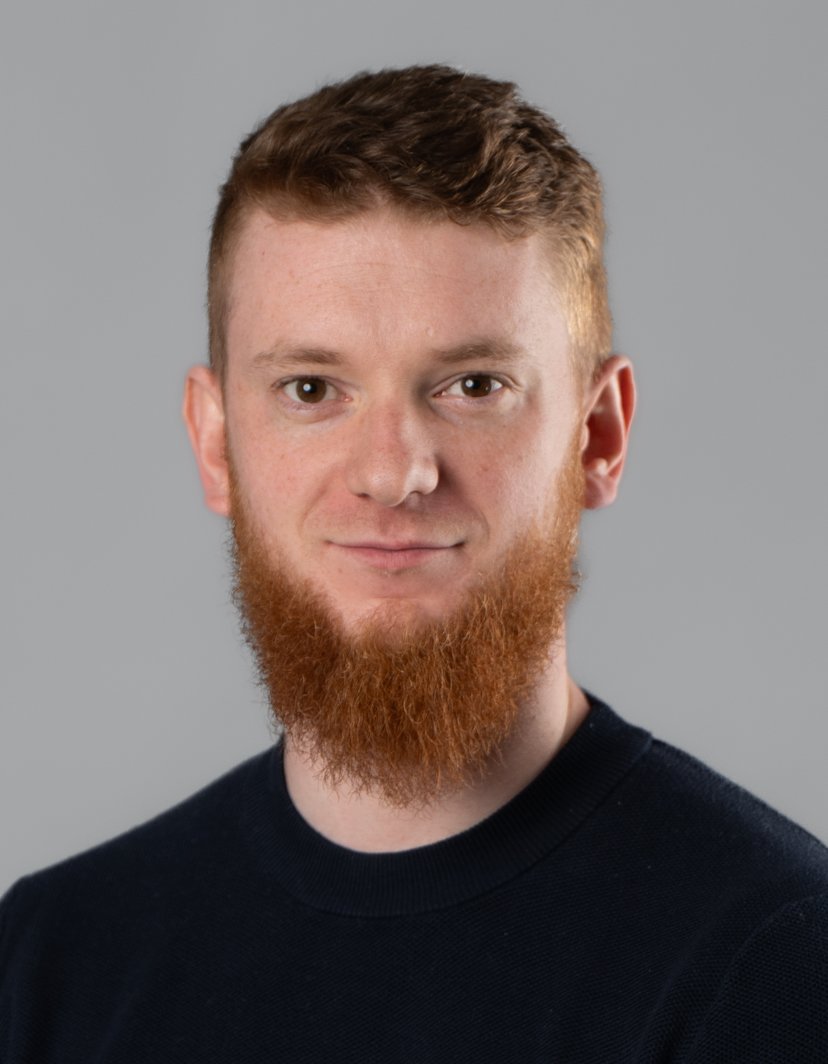}}]{Frantisek Rusnak}
František Rusnák received the B.S. and M.S. degrees in automation and measurement from Brno University of Technology, Brno, Czech Republic in 2021 and 2023, respectively. He is currently pursuing a Ph.D. degree in cybernetics, control and measurement at the same institution. His research focuses on the analysis and modeling of connectivity and time synchronization in communicating nodes. He is currently involved in a research project dedicated to the development of an artificial intelligence controlled guided drone.
\end{IEEEbiography}

\begin{IEEEbiography}[{\includegraphics[width=1in,height=1.25in,clip,keepaspectratio]{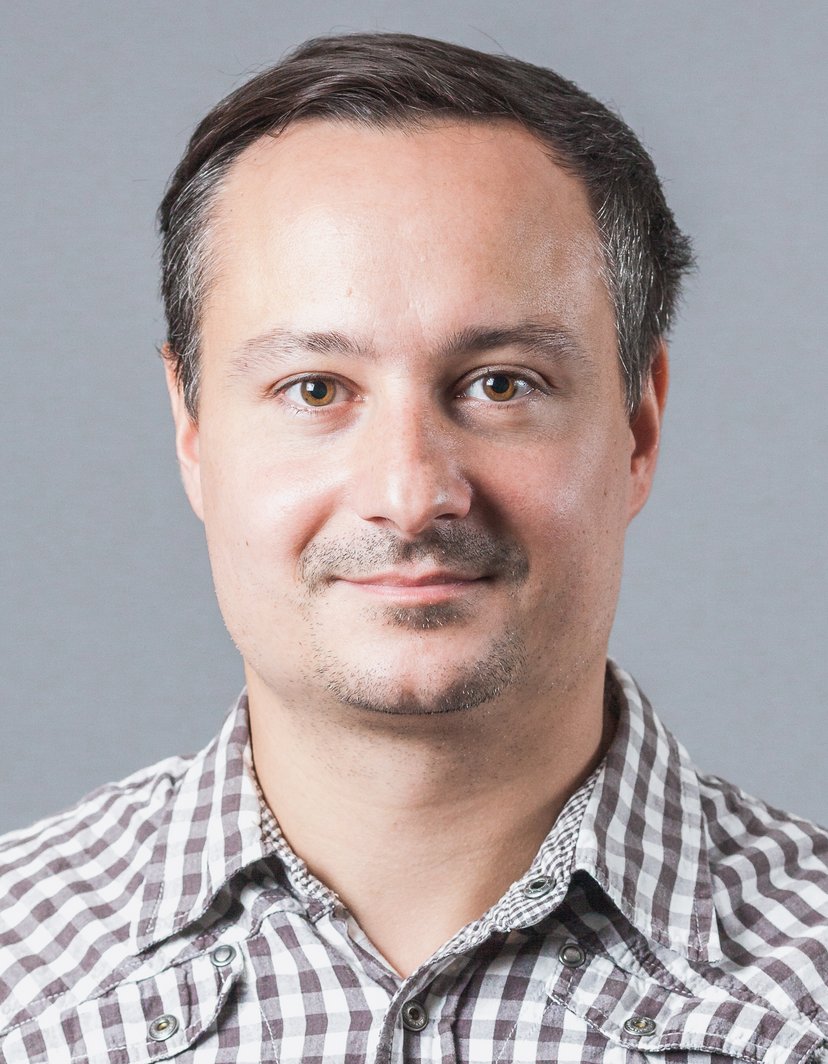}}]{Ondrej Cech}
Ing. Ondřej Čech, Ph.D., is an Ass. Prof. at the Department of Electrotechnology (UETE), Faculty of Electrical Engineering and Communication, Brno University of Technology (BUT). He obtained his Ph.D. degree at the Brno University of Technology, Faculty of Electrical Engineering and Communication, specializing in Electrotechnology. His research activities encompass both fundamental research in the field of electrochemical energy storage, structural characterization of novel materials for Li-ion and Li-S batteries and applied research in areas involving monitoring systems. 
\end{IEEEbiography}

\begin{IEEEbiography}[{\includegraphics[width=1in,height=1.25in,clip,keepaspectratio]{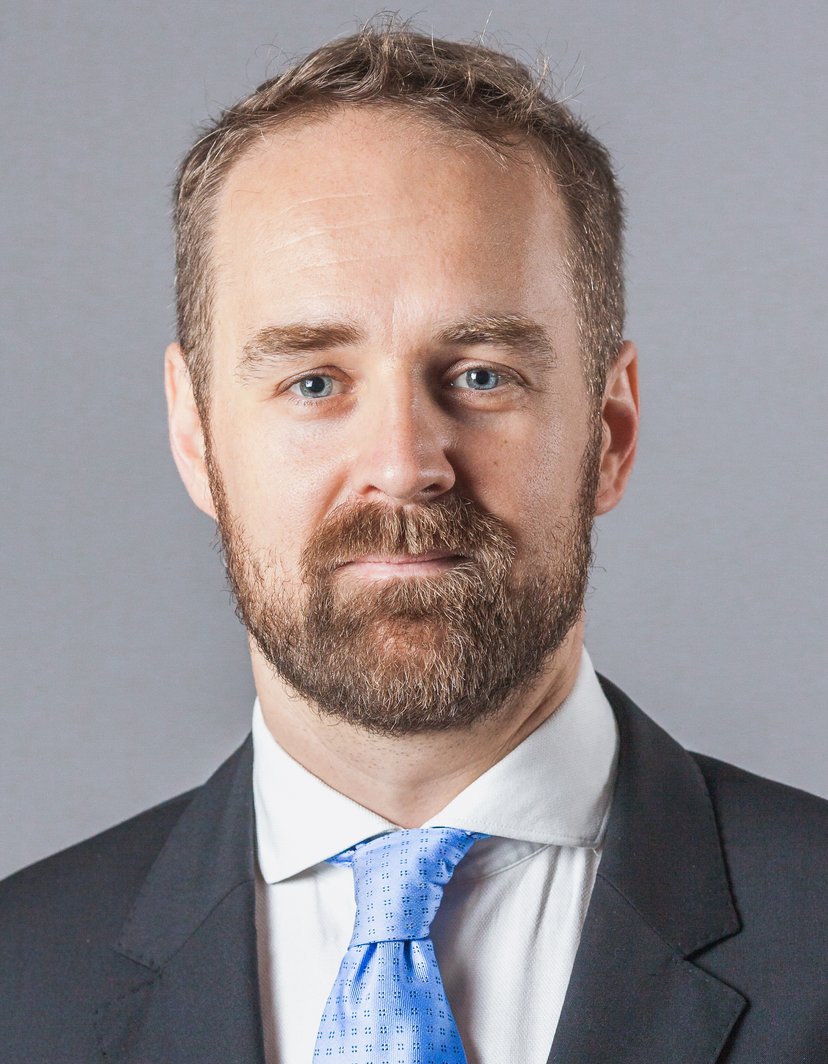}}]{Karel Horak}
Dr. Horak is Ass. Prof. in the Department of Control and Instrumentation at Brno University of Technology. His expertise is of computer vision, optics, machine learning and signal processing. He obtained his master and PhD degrees in cybernetics in 2004 and 2008 respectively and he is the head of Machine Vision Group since 2011. He obtained a Top10 excellence award at BUT and Czech Engineering Academy award for example.  He has been as a research fellowship at Technischen Universität Wien and at Austrian Institute of Technology in Austria and at Lappeenranta-Lahti University of Technology in Finland.
\end{IEEEbiography}




\end{document}